\documentclass[letterpaper, 10 pt, conference]{ieeeconf}  

\IEEEoverridecommandlockouts                              

\usepackage[T1]{fontenc}

\newcommand{\vertiii}[1]{{\left\vert\kern-0.25ex\left\vert\kern-0.25ex\left\vert #1 
    \right\vert\kern-0.25ex\right\vert\kern-0.25ex\right\vert}}

\usepackage{graphics} 
\usepackage{epsfig} 
\usepackage{times} 
\usepackage{amsmath} 
\usepackage{amssymb}  
\let\labelindent\relax
\usepackage{enumitem}
\usepackage{multicol}
\usepackage{multirow}
\usepackage{graphicx}
\usepackage{xcolor}
\usepackage{float}
\usepackage{subfiles}
\usepackage{tabularx}
\usepackage{booktabs}
\usepackage{arydshln}
\usepackage[font=small]{caption}
\usepackage{bm}
\usepackage{subcaption}
\usepackage{hyperref}
\usepackage[font=footnotesize,labelfont=bf]{caption}
\newtheorem{theorem}{Theorem}[section]

\newtheorem{definition}[theorem]{Definition}
\newtheorem{remark}[theorem]{Remark}

\title{\LARGE \bf
Barrier Functions Inspired Reward Shaping for \\ Reinforcement Learning 
}

\author{
Nilaksh$^{1*}$,
Abhishek Ranjan$^{2*}$,
Shreenabh Agrawal$^{2*}$,
Aayush Jain$^{1}$,
\thanks{This work is supported in part by the Google Research Grant, the ARTPARK, and the SERB Grants SRG/2022/001807 and CRG/2021/008115.}
\thanks{$^{1}$Indian Institute of Technology (IIT), Kharagpur
	    {\tt\small \{nilaksh404, aayushjain\}@kgpian.iitkgp.ac.in}
}
\thanks{$^{2}$Indian Institute of Science (IISc), Bangalore
	    {\tt\small \{abhishekr, shreenabhm\}@iisc.ac.in}
}
Pushpak Jagtap$^{3}$,
\thanks{$^{3}$RBCCPS, IISc - {\tt\small pushpak@iisc.ac.in}}
Shishir Kolathaya$^{4}$
\thanks{$^{4}$CSA \& RBCCPS, IISc - {\tt\small shishirk@iisc.ac.in}}
\thanks{$^{*}$Equal Contribution}}

\begin{document}

\maketitle
\thispagestyle{empty}
\pagestyle{empty}

\begin{abstract}

Reinforcement Learning (RL) has progressed from simple control tasks to complex real-world challenges with large state spaces. While RL excels in these tasks, training time remains a limitation. Reward shaping is a popular solution, but existing methods often rely on value functions, which face scalability issues. This paper presents a novel safety-oriented reward-shaping framework inspired by barrier functions, offering simplicity and ease of implementation across various environments and tasks. To evaluate the effectiveness of the proposed reward formulations, we conduct simulation experiments on CartPole, Ant, and Humanoid environments, along with real-world deployment on the Unitree Go1 quadruped robot. Our results demonstrate that our method leads to 1.4-2.8 times faster convergence and as low as 50-60\% actuation effort compared to the vanilla reward. In a sim-to-real experiment with the Go1 robot, we demonstrated better control and dynamics of the bot with our reward framework. We have open-sourced our code at \url{https://github.com/Safe-RL-IISc/barrier_shaping}.
\end{abstract}

\section{INTRODUCTION}

Reinforcement Learning (RL) has demonstrated substantial success in various domains, including gaming (e.g., Minecraft \cite{hafner2023mastering} and Atari \cite{mnih2015human}), language model optimization (e.g., Sparrow \cite{glaese2022improving} and InstructGPT \cite{ouyang2022training}), and robotics \cite{arimoto1984bettering,doi:10.1177/0278364913495721}. However, all of these methods including RLHF \cite{christiano2017deep}, which addresses RL's sample efficiency issues in other domains, can be costly when applied to robotics, including sim-to-real transfer scenarios. Despite having its limitations, reward shaping provides a simpler, more accessible and efficient alternative to address these issues even now. Well-crafted reward functions guide the agent's behaviour towards desired outcomes, facilitating successful learning of the intended task. Previous works (\cite{10.1145/1273496.1273572}, \cite{4670492}) have established reward-shaping to accelerate algorithm convergence. Potential-based reward shaping \cite{ng1999policy} is a well-known work that suggests adding a potential function term initialized to the value function for reward-shaping. However, it needs a good estimate of the value function, which is challenging in the case of sparse reward models and in complex environments due to dimensionality \cite{dong2020principled}, \cite{barto2003recent}, \cite{bellman1966dynamic}.
\begin{figure}[!t]
    \centering
    \begin{subfigure}[b]{0.50\textwidth}
        \centering
        \includegraphics[width=0.95\linewidth]{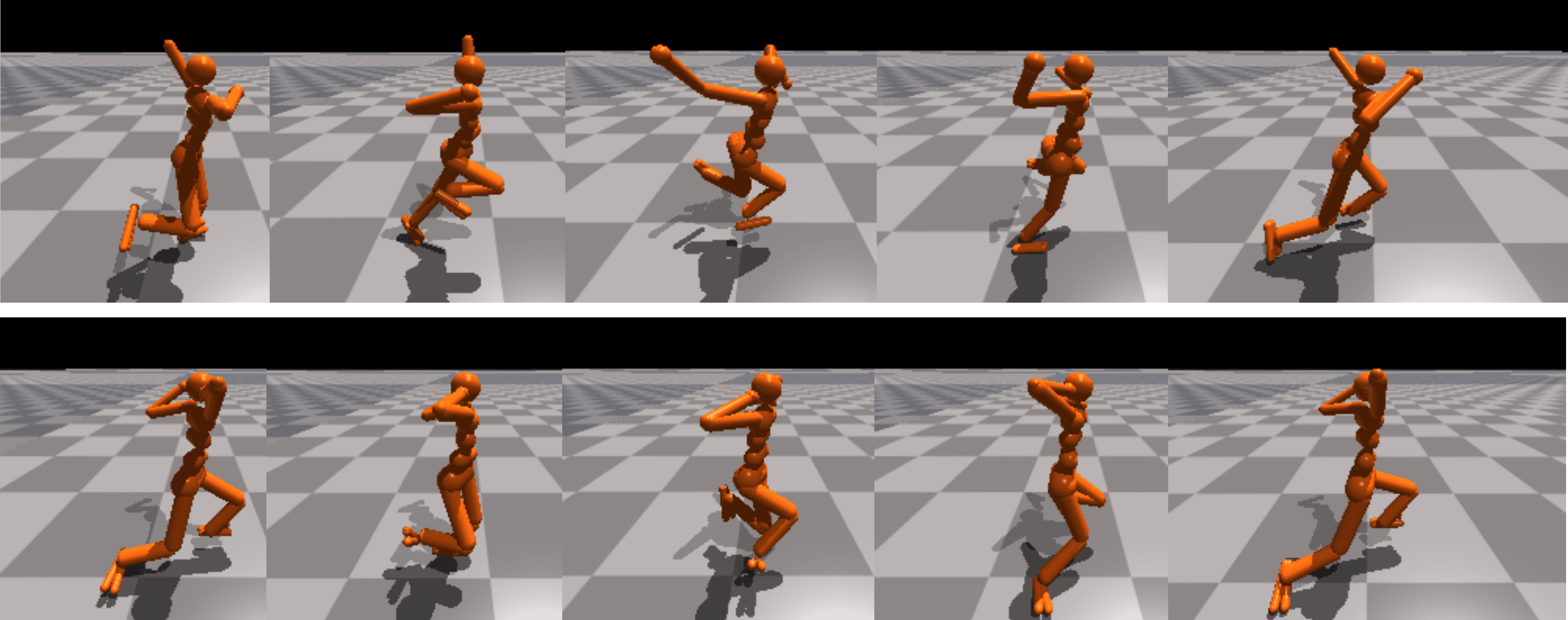}
                \label{fig:ant_run_exp}
    \end{subfigure}
    \caption{\small An example run of Humanoid with vanilla reward (Top) and the exponential barrier reward (Bottom), trained for the same number of time steps. Unlike the vanilla reward, barrier function-based reward leads to more natural and less wasteful movements.}\label{fig:running}
    \vspace{-0.5cm}
\end{figure}

With a view to improve upon existing works on reward shaping methods, we introduce Barrier Function (BF) inspired reward shaping, a simple, safety-focused framework that enhances training efficiency and safety. Our approach uses BFs to supplement the base reward in an environment. Intuitively, when a BF is positive, the system's state is within the safe region. Accordingly, we can construct an inequality by using the derivative of the barrier function, which is in turn encoded in the form of a reward. This reward-shaping term encourages the RL agent's states to not only remain within the safe zone, but also ensure that undesirable behaviors are avoided at the limits. In this paper, we propose two BF-based reward formulations: the \textit{exponential barrier} and the \textit{quadratic barrier}. We assess our framework across various environments, from CartPole\cite{c4} to Humanoid\cite{c2}. Additionally, our approach improves the walking performance of agents, as evident in Fig \ref{fig:running}. We also demonstrate sim-to-real transferability by applying our framework to the Unitree Go1 robot \cite{unitree2022website}.
The key highlights of the proposed framework are:
\begin{itemize}
    \item A safety-oriented, intuitive and easy-to-implement barrier function-inspired reward shaping framework.
    \item The framework leads to faster convergence towards the goal and efficient state exploration by enforcing the system within the safe set. 
    \item It leads to lesser energy expenditure as the barrier function constrains the states within desired limits, thus avoiding extreme actions.
\end{itemize}

\section{Related Work}
\textit{Reward Shaping :} Positive linear transformation \cite{c7}, \cite{c9}, \cite{c10}, leverages the principles of utility theory to enhance agent performance by adding rewards for state transitions as the difference between the values of the arbitrary potential functions applied to the respective states. In \cite{c15}, a Bayesian approach to reward shaping is presented, which incorporates prior beliefs and adapts with experience. While effective, this method is model-based and may not generalize well across different environments. Signal temporal logic, explored in \cite{c18} and \cite{saxena2023funnel}, can serve as a formal specification for reward formulation but is also model-dependent. The safety and stability of the agent and environment remain unaddressed. To the best of our knowledge, for the first time, we explore the agent's safety and stability using reward shaping that ensures faster convergence along with less energy consumption during training and testing.

It is worth mentioning that, in our formulation, we omit energy terms from the shaped reward as they often clash with the environment's goals. For example, in CartPole, including an energy term would prioritize having the pendulum vertically down, which isn't the intended objective. Adjusting energy terms requires extra tuning, complicating things for larger robot models. Similarly, alternative reward-shaping approaches like angle limits and velocity limits may lead to undesirable behavior, keeping the agent within bounds but failing to achieve the main objective.  Intuitively, there is a relationship between the positions and velocities that must be respected at the limits. Violation of this can restrict exploration and learning, hindering effective policy development. The CBF based constraints encode this relationship, thereby ensuring that the position-velocity values are not in conflict. 

\textit{Safety :} \cite{Cheng_Orosz_Murray_Burdick_2019} introduces a framework that combines model-free reinforcement learning with model-based CBF controllers and learned system dynamics to ensure safety during exploration. Since it is model-based, a good knowledge of the system dynamics is required. \cite{yu2022reachability} proposes using reachability constraints to expand the feasible set, resulting in a less conservative policy compared to CBF-based approaches. \cite{zhao2020learning} explores a more data-driven approach using two neural networks to learn the controller and safety barrier certificate simultaneously achieving a verification-in-the-loop synthesis, but requires formal verification (using SMT solvers) to ensure constraint satisfaction. Lastly, \cite{c14} discusses learning to restructure an MDP reward function for accelerated reinforcement learning, but it can be computationally intensive. Tan et al. \cite{tan2023your} link CBF to a value function to enforce verifiable safety. Our approach aligns with the idea of leveraging model-based constraints for improved performance, but it doesn't require learning system dynamics, making it more generalizable. We also allow for adjustable constraint levels (soft or hard) to prioritize returns or constraint satisfaction. Furthermore, we eliminate the need for formal verification, as our formulation inherently satisfies constraints.
\section{Preliminary}
\subsection{Reinforcement Learning}
Reinforcement learning (RL) can be described as a discounted Markov decision process (MDP), defined by the tuple $\mathcal{M}$ = ($\mathcal{S, A, P,}$ $\emph{r}$, $\gamma$), where $\mathcal{S}$ is a set of states, $\mathcal{A}$ is a set of actions, $\mathcal{P}: \mathcal S \times A \rightarrow \mathcal{S}$ is the deterministic state transition function, $\emph{r}: \mathcal{S} \times A \rightarrow \mathbb{R}$ is the reward function, and $\gamma \in$ (0,1) is the discount factor. In RL, the goal of the learning algorithm is to converge on a policy $\pi$: $\mathcal{S} \rightarrow \mathcal{A}$ that maximizes the total (discounted) reward after performing actions on an MDP, i.e., the objective is to maximize $\Sigma_0^\infty \gamma^t r_t$, where $r_t$ is the output of the reward function $r$ for the sample at instance $t$. Since a policy maps a state to an action, the value of a policy is evaluated according to the discounted cumulative reward. Given this MDP, we are interested in shaping the rewards by using barrier functions, described in the next section. 

\subsection{Barrier Functions}
For practical applications, we often want the system state $s$ to stay within a safe region, denoted as a set $\mathcal{C}$. The set $\mathcal{C}$ is defined as the \textit{super-level set} of a continuously differentiable function $h:\mathcal{S}\subseteq \mathbb{R}^n \rightarrow \mathbb{R}$ satisfying,
\begin{align}
\label{eq:setc1}
	\mathcal{C}                        & = \{ s \in \mathcal{S} : h(s) \geq 0\} \\
\label{eq:setc2}
	\partial\mathcal{C}                & = \{ s \in \mathcal{S}: h(s) = 0\}\\
\label{eq:setc3}
	\text{Int}\left(\mathcal{C}\right) & = \{ s \in \mathcal{S}: h(s) > 0\},
\end{align}
where $\text{Int}\left(\mathcal{C}\right)$ and $\partial\mathcal{C}$ denote the interior and
the boundary of the set $\mathcal{C}$, respectively.
It is assumed that $\text{Int}\left(\mathcal{C}\right)$ is non-empty and $\mathcal{C}$ has no isolated points, i.e. $\text{Int}\left(\mathcal{C}\right) \neq \phi$ and $\overline{\text{Int}\left(\mathcal{C}\right)} = \mathcal{C}$. The set $\mathcal{C}$ is said to be forward invariant (safe) if $\forall \: s(0) \in \mathcal{C} \implies s(t) \in \mathcal{C} \;\;\; \forall t \geq 0$.
We can mathematically verify the safety of the set $\mathcal{C}$ by establishing the existence of a barrier function.
We have the following definition of a barrier function (BF) from \cite{c20}.
\vspace{2mm}
\begin{definition}
{\it
\label{definition: BF definition}
Given the set $\mathcal{C}$ defined by \eqref{eq:setc1}-\eqref{eq:setc3}, 
the function $h$ is called the  barrier function (BF) defined on the set $\mathcal{S}$ if there exists an extended \textit{class} $\mathcal{K}$ function $\kappa$ such that for all $s \in \mathcal{S}$:
\begin{equation}\label{eq:bfineq}
\begin{aligned}
     \dot{h}\left(s, \dot s \right) \! + \kappa\left(h(s)\right) \! \geq \! 0.
\end{aligned}
\end{equation}}
\end{definition}
\vspace{2mm}

Here $\kappa : \left(-\infty, \infty\right) \rightarrow (-\infty, \infty)$ is a strictly increasing continuous function, with $\kappa$(0) = 0. $\kappa$ is widely called an extended class $\mathcal{K}$ function.
Note that $\dot{h}(s,\dot{s}):=\frac{\partial h}{ \partial s}\dot s$, where $\dot s$ is the time derivative of $s$. Even if the MDP is in discrete time, $\dot s$, and consequently $\dot h$, can be calculated approximately as $\dot h  \approx \frac{\partial h}{ \partial s} (s_t - s_{t-1})/ \Delta t$, where $s_t, s_{t-1}$ are the samples of the states obtained at time steps $t$ and $t-1$, and $\Delta t$ is the time interval between two samples. It is worth mentioning that the classical definition has the actions or controls $u$ as one of the arguments along with model in \eqref{eq:bfineq}, thereby making it a {\it control} barrier function (CBF). For this paper, we avoid the use of inputs and make estimates of the derivative of $h$ by using $\dot s$. This allows us to use barrier functions in a model-free way and is sufficient for the presented work as our focus is on reward shaping. 

If we are able to restrict the states of the system $s,\dot s$ in such a way that the inequality \eqref{eq:bfineq} is satisfied, then we know that the set $\mathcal{C}$ is forward invariant (safe). We can use this idea to shape the reward functions in such a way that any violation of safety causes a loss of reward. We will formally show the reward-shaping methodology in the next section. 

\setlength{\abovedisplayskip}{2pt}

\section{Reward Shaping Methodology}
\subsection{Reward Shaping using Barrier Functions}
Having described BFs and their associated formal results, we now discuss BF-inspired reward shaping in the context of RL. Reward shaping is a method for engineering a reward function to provide more frequent feedback on appropriate behaviours. To illustrate our framework, we propose the shaped reward $r'$ in \eqref{eq:shaping}, which we obtain by adding a term to the traditional vanilla reward $r$.
\begin{align}
    r'\left(s, \dot s\right) = r(s) + \underbrace{r^\text{BF}\left(s, \dot s\right)}_{\text {additional reward }},
    \label{eq:shaping}
\end{align}
\vspace{-0.3cm}

$r^\text{BF}(s,\dot s)$ is our barrier function inspired reward shaping term. We emphasise that the new reward $r'$ depends on $\dot{s}$.  
From Definition \ref{definition: BF definition}, $\mathcal{C}$ is forward invariant if and only if there exists a barrier function $h$ such that it satisfies \eqref{eq:bfineq}. Thus we define $r^\text{BF}$ as 
\begin{align}
\label{eq:shaping-term1}
    r^\text{BF}(s, \dot s)  = \dot h(s,\dot s) + \gamma h(s).
    \vspace{-0.25cm}
\end{align}
In order to satisfy \eqref{eq:bfineq} we have taken the extended class $\mathcal{K}$ function $\kappa$ as $\kappa(m) = \gamma m$, $\gamma \in \mathbb{R}_{>0}$. 
This gives our shaping term $r^\text{BF}$ the desirable property of positively rewarding the agent when it is in a set of safe states $\mathcal{C}$ while negatively rewarding otherwise. 

The BF $h$: $\mathcal{S} \rightarrow \mathbb{R}$ is chosen to be a suitable function that constrains specific quantities in $\mathcal{S}$, such that it would lead to desirable properties like actuation safety and training efficiency. The following section provides specific examples of BF-based reward shaping.

\begin{remark}
Since, in RL task, our goal is to find a policy $\pi:\mathcal{S}\rightarrow\mathcal{A}$ that maximizes total (discounted) reward, as we maximise the proposed reward \eqref{eq:shaping} it will also enforce the condition \eqref{eq:bfineq} in Definition \ref{definition: BF definition} and thus implies safe execution of the task. However, it is essential to note that this approach does not provide a safety guarantee and does not imply safety during training.
\end{remark}
\subsection{Barrier Function Formulation}
\begin{figure}[h!]
    \centering
    \vspace{-0.2cm}
        \includegraphics[width=0.8\columnwidth]{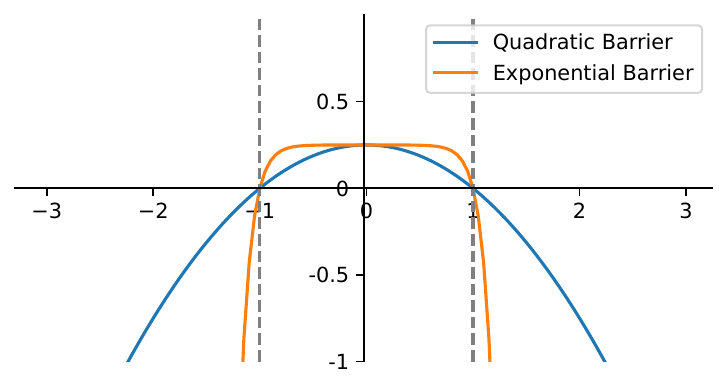}
    \vspace{-0.2cm}
    \caption{Plots illustrating the proposed barrier functions \eqref{eq:quad-cbf}-\eqref{eq:exp-cbf}. Dashed lines represent the constraint limits (-1,1).}
        \label{fig:barriers}
    \vspace{-0.3cm}
\end{figure}
In an RL task, some state variables should ideally lie between a safe range of values. Violating these bounds can result in undesirable behaviours. To constrain them within their safe bounds, we must use an appropriate barrier function, $h(s, \bm{\delta}) \equiv h(s)$ parameterized by $\bm{\delta}$.   
We propose two barrier functions: a quadratic function $h_\text{quad}$ \eqref{eq:quad-cbf} and an exponential function $h_\text{exp}$ \eqref{eq:exp-cbf}.
    \begin{align}
            \label{eq:quad-cbf}
            h_\text{quad}(s, \bm{\delta}) & = \sum_{l \in \mathcal{L}}^{ }\delta_a\left(s_{l}-s^\text{max}_{l}\right)\left(s^\text{min}_{l}-s_l\right) \\
            \label{eq:exp-cbf}
            h_\text{exp}(s, \bm{\delta}) & =\sum_{l\in \mathcal{L}}^{ }\delta_a\left[1-\left(e^{\delta_b\left(s_{l}-s^\text{max}_{l}\right)}+e^{\delta_b\left(s^\text{min}_{l}-s_{l}\right)}\right)\right]
    \end{align}
where $\bm{\delta} = [\delta_a, \delta_b]$, $\bm{\delta} \in \mathbb{R}^2_{>0}$ is the vector parameter, and $\mathcal{L} = \{l_0, l_1, l_2, \dots \}$ is the set of indices for the elements of the state variables of the model whose values we want to constrain. In particular, $\{s_{l}\}$ is $l^{th}$ element of state vector $s$, which is the value of the state variable upon which we wish to enforce the bounds ($s^\text{min}_{l}$, $s^\text{max}_{l}$). The choice of $\bm{\delta}$ and the bounds ($s^\text{min}_{l}$, $s^\text{max}_{l}$) is elaborated in the following section.

Since $h(s, \bm{\delta})$ is positive for an $l\in\mathcal{L}$ only when $s_l\in(s_l^\text{min}, s_l^\text{max})$, and negative otherwise, $h(s, \bm{\delta})$ qualifies as a barrier function and $r^\text{BF}$ can be computed by Eq. \eqref{eq:shaping-term1}.

For appropriate values of $\bm{\delta}$, $h_\text{exp}$ is flat within the bounds and tapers down sharply outside them, which allows for better exploration while staying within the bounds. In contrast, the shape of $h_\text{quad}$ makes it more suitable for tasks that require $s_l$ to be constrained near the central values (Fig. \ref{fig:barriers}).
\vspace{-0.3cm}
\begin{figure}[h!]
    \centering
        \includegraphics[width=0.6\columnwidth]{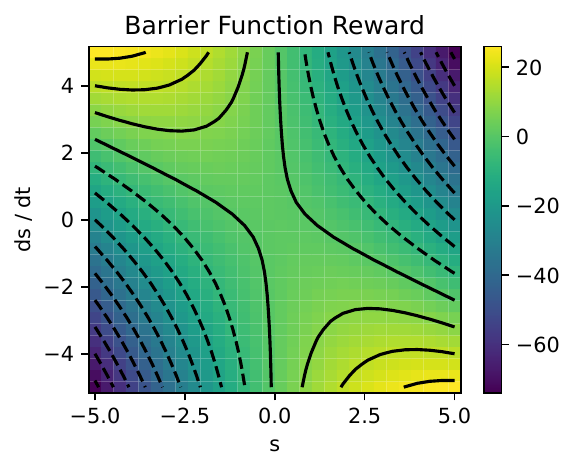}
    \caption{Density plot of the quadratic BF reward, $r^\text{BF}_\text{quad}$ constructed using \eqref{eq:shaping-term1} and \eqref{eq:quad-cbf} with (-1,1) as the bounds on $s$. Notice that the reward depends on $s$ and $\dot s$. The solid and dotted contour lines correspond to positive and negative values of the reward, respectively.}
        \label{fig:BF2D}
\end{figure}
\vspace{-0.3cm}

The reward $r^\text{BF}$ is a function of both $s$ and $\dot s$. Thus, it jointly constrains them depending on their values (Fig. \ref{fig:BF2D}). To get a more intuitive understanding, consider the following,
\begin{align}
    \begin{split}
        r^\text{BF}(s, \dot s)  &= \dot h(s,\dot s) + \gamma h(s) \\
                                &= \frac{\partial h(s)}{\partial s} \frac{d s}{d t}+ \gamma h(s) =\frac{\partial h(s)}{\partial s} \dot s + \gamma h(s)
    \end{split}
    \label{eq:bf-intuit1}
\end{align}
\begin{figure*}[t]
 \begin{minipage}[b]{.32\textwidth}
  \centering
  \includegraphics[width=1.12\textwidth]{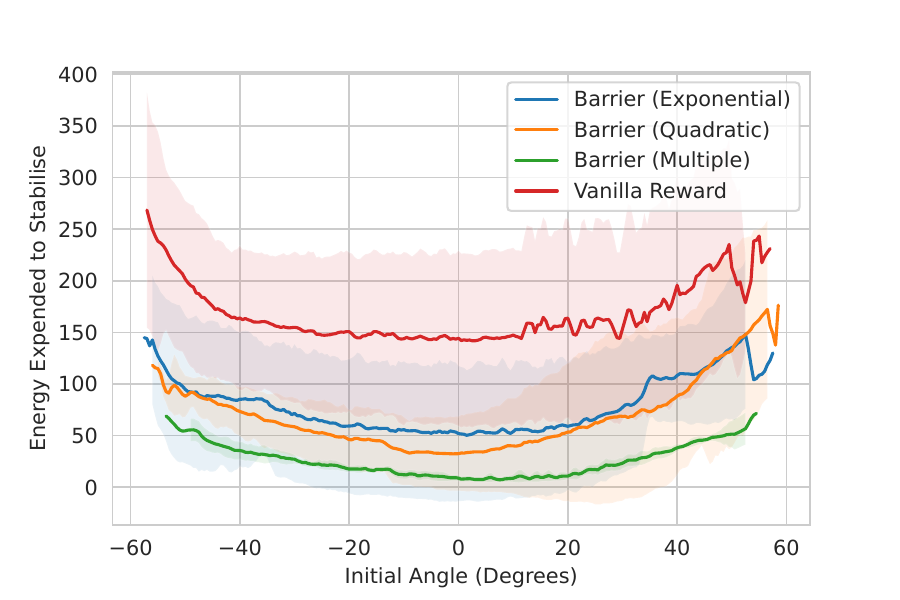}
  \caption{Energy expended to stabilize initial angles for each reward formulation in the cartpole environment}
  \label{fig:cp1}
  \end{minipage}%
  \hfill
  \begin{minipage}[b]{.66\textwidth}
  {\includegraphics[width = 0.45\textwidth]{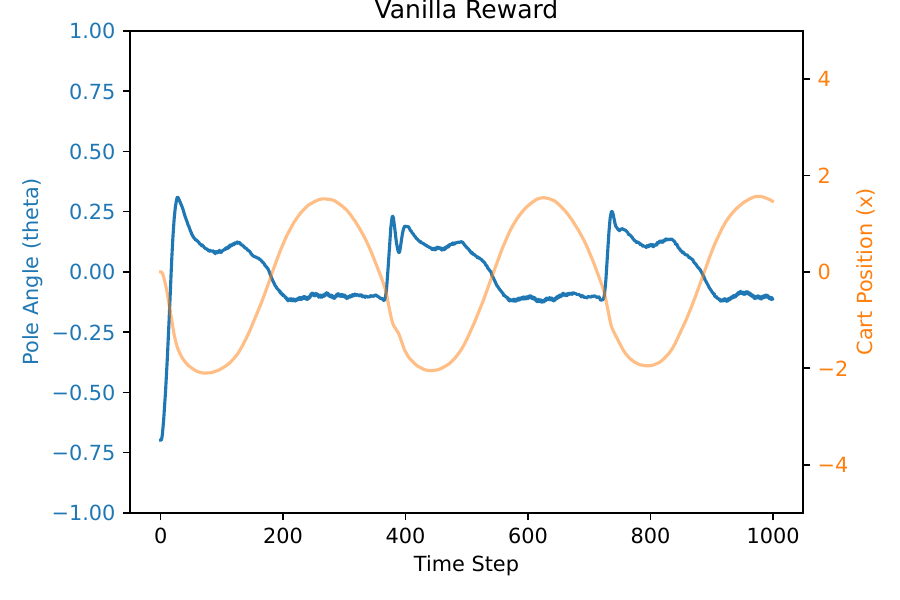}}
  {\includegraphics[width = 0.45\textwidth]{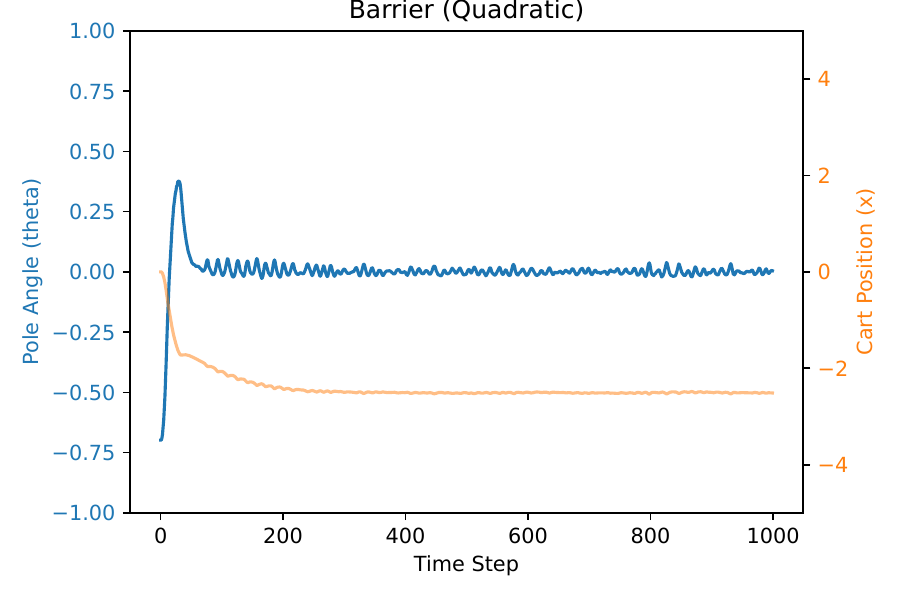}}  
  \caption{$\theta_p$ and x-position vs time-step graph showing control flow to stabilize -40$^\circ$ pole angle in the cart pole environment. The policy clearly shows that the vanilla policy struggles to stabilize the pole angle close to zero, while the quadratic policy accomplishes this in a few time-steps.\\}
            \label{fig:Cartpole2}
  \end{minipage} 
  \vspace{-0.7cm}
\end{figure*}
Taking the partial derivative of $r^\text{BF}$ with respect to $\dot s$ gives,

\begin{align}
    \begin{split}
        \frac{\partial r^\text{BF}(s, \dot s)}{\partial \dot s}  &= \frac{\partial}{\partial \dot s} \frac{\partial h(s)}{\partial s} \dot s + \frac{\partial}{\partial \dot s}\left(\gamma h(s)\right) 
                                = \frac{\partial h(s)}{\partial s} = h'(s)
    \end{split}
    \label{eq:bf-intuit2}
\end{align}
Now, if the barrier function $h(s)$ is taken to be a concave function, then there exists an $s_0 \in (s^\text{min}, s^\text{max})$ such that $h'(s) > 0$ for all $s < s_0$ and negative otherwise. Thus, from \eqref{eq:bf-intuit2} we can infer that when $s < s_0$ i.e. it is near the lower bound, $(\partial / \partial \dot s)r^\text{BF} = h'(s)$ is positive. As an RL policy wants to maximize $r^\text{BF}$, it promotes $\dot s$ to be positive. Given that $\dot s$ denotes the rate of change of $s$ over time, a positive $\dot s$ causes $s$ to increase, moving it away from the lower bound $s^\text{min}$. Conversely, the opposite effect applies when $s$ is near $s^\text{max}$. In both cases, $s$ is promoted to move towards $s_0$.

Thus, $r^\text{BF}$ works with both $s$ and $\dot s$ to promote safety. This effect is lacking with other reward shaping methods that constrain only $s$ within some bounds for safety. It can be verified that both of our proposed barriers $h_\text{quad}$ and $h_\text{exp}$ are concave functions with $s_0 = (s^\text{max} + s^\text{min})/2$. 

\section{Simulation Experiments}
We experimentally evaluate our BF-based reward-shaping formulation on OpenAI Gym's Cartpole \cite{c1}, and MuJoCo environments like Half-Cheetah \cite{c3}, Humanoid \cite{c2}, and Ant \cite{c4}. We used the Twin Delayed DDPG (TD3) \cite{fujimoto2018addressing} algorithm with two variants of the BF-based reward shaping: $r^\text{BF}_\text{quad}$ \eqref{eq:quad-cbf} and $r^\text{BF}_\text{exp}$ \eqref{eq:exp-cbf}, which yield the $\pi^\text{BF}_\text{quad}$ and $\pi^\text{BF}_\text{exp}$ policies respectively. The environment's \textit{vanilla} reward is used as the baseline. All experiments were performed on a Ryzen Threadripper CPU, 64GB RAM, and an NVIDIA RTX 3080.
\subsection{Cartpole}
\begin{figure*}[t]
        \centering
        \begin{minipage}[b]{0.33\textwidth}
            \centering
            \includegraphics[width=\textwidth]{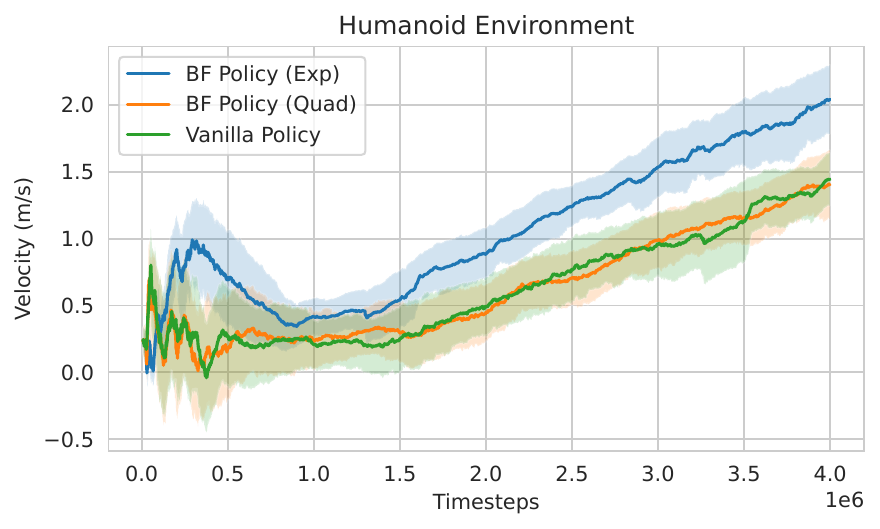}
        \end{minipage}
        \hspace{-1em}
        \begin{minipage}[b]{0.33\textwidth}
            \centering
            \includegraphics[width=\textwidth]{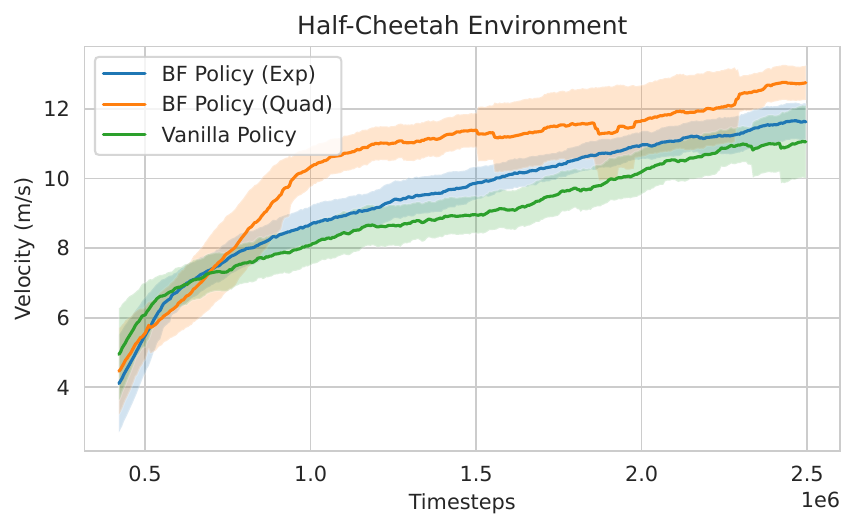}
        \end{minipage}
        \hspace{-1em}
        \begin{minipage}[b]{0.33\textwidth}
            \centering
            \includegraphics[width=\textwidth]{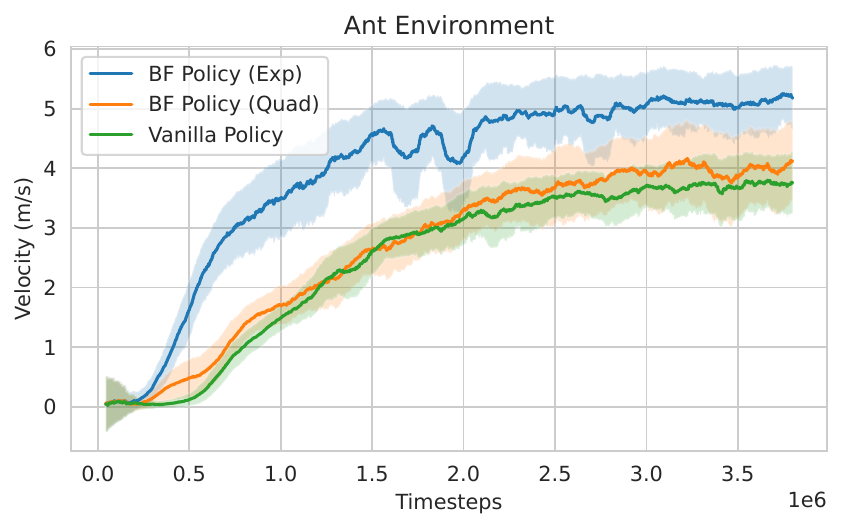}
        \end{minipage}
        \caption{The plots for each MuJoCo walker environment, averaged for ten random seeds, show the episodic velocity for each training time step. Since these environments aim to achieve as high a velocity as possible, these plots provide a good metric to judge the training speed.}
        \label{fig:vel-train}
        \vspace{0.1cm}
        
        \begin{minipage}[b]{0.33\textwidth}
            \centering
            \includegraphics[width=\textwidth]{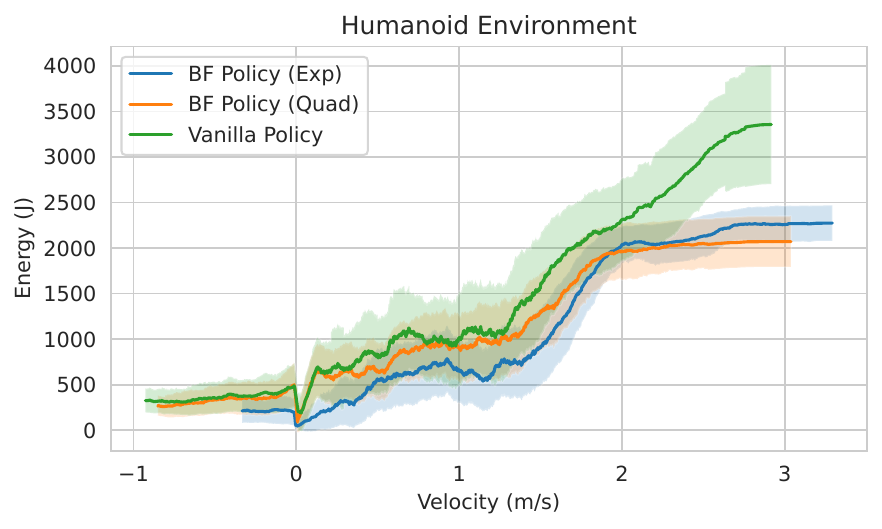}
        \end{minipage}
        \hspace{-1em}
        \begin{minipage}[b]{0.33\textwidth}
            \centering
            \includegraphics[width=\textwidth]{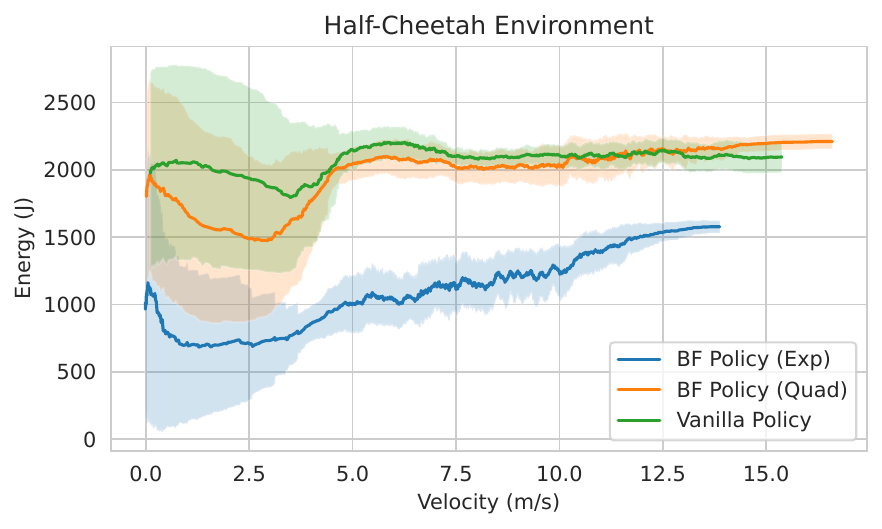}
        \end{minipage}
        \hspace{-1em}
        \begin{minipage}[b]{0.33\textwidth}
            \centering
            \includegraphics[width=\textwidth]{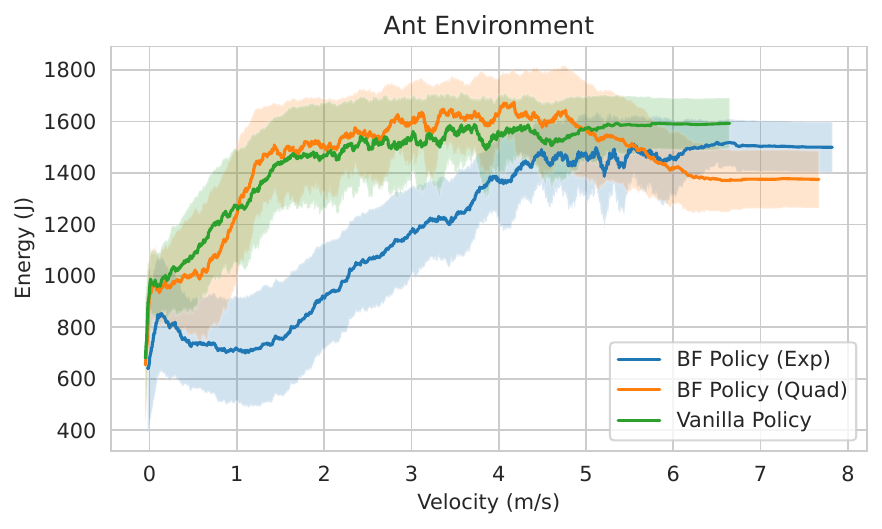}
        \end{minipage}
        \caption{The episodic energy (in joules) spent by the agent for achieving a particular velocity, averaged over ten random seeds. We see that the policy trained with exponential BF-based reward shaping performs the best across all the environments, consuming the least energy.}
        \label{fig:enrg-vel}
    \end{figure*}
    \begin{table*}[t]
\vspace{1mm}
\centering
\begin{tabular}{@{}ccccccccc@{}}
\toprule
\multirow{2}{*}{Policy}             & \multicolumn{2}{c}{Humanoid}                     & \multicolumn{2}{c}{Half-Cheetah}               & \multicolumn{2}{c}{Ant}                         \\ \cmidrule(l){2-9} 
                                    & Rel. Actuation Coeff.                    & Rel. Time             & Rel. Actuation Coeff.                 & Rel. Time             & Rel. Actuation Coeff.                  & Rel. Time             \\ \cmidrule(r){1-1}
Vanilla                             & 1.00         & 1.00                  & 1.00        & 1.00                  & 1.00         & 1.00                  \\
\hdashline\noalign{\vskip 0.5ex}
Quadratic BF                       & 0.66 $\pm$ 0.08         & 0.89 $\pm$ 0.12          & 0.97 $\pm$ 0.08         & \textbf{0.68 $\pm$0.03} & \textbf{0.62 $\pm$0.05} & 0.76 $\pm$ 0.15          \\
Exponential BF                     & \textbf{0.49 $\pm$0.03} & \textbf{0.64 $\pm$0.06} & \textbf{0.92 $\pm$0.05} & 0.96 $\pm$ 0.02          & 0.63 $\pm$ 0.06          & \textbf{0.36 $\pm$0.05} \\ \bottomrule
\end{tabular}
\caption{Experiment results for different environments and reward formulations. The \textit{Relative Actuation Coefficient} field is the episodic energy expended in actuation (in joules) divided by the squared velocity (m$^2$s$^{-2}$) relative to vanilla reward. The result is averaged over 100 best runs. \textit{Relative Time} is the relative training time steps taken by a reward formulation to achieve the same maximum velocity as the Vanilla reward (averaged over 10 random seeds).}
\vspace{-0.5cm}
\label{tab:main-res}
\end{table*}

\noindent \textbf{Reward Shaping:} The state-space contains the pole angle $\theta_p$, pole angular velocity $\omega_p$, cart's position $x_c$, and velocity $v_c$. The task consists of balancing a pole attached to a moving cart, i.e. we want $ \theta_p = 0$. The vanilla reward is $r$ = +1 for every step taken. We define $(\theta_p^\text{min}, \theta_p^\text{max}) \equiv (s_l^\text{min}, s_l^\text{max})$ as the pole angle's desired threshold range for the goal ($\theta_p$ = 0) around which we want the pendulum to stabilize, hence $h = h(\theta_p, \bm{\delta})$. This bound is taken as the maximum angle deviation $\phi$ from which the pole can balance itself. According \eqref{eq:shaping}-\eqref{eq:shaping-term1}, the quadratic BF \eqref{eq:quad-cbf} shaped reward is given as, (taking $\gamma$ = 1, $\theta_p^\text{max} = -\theta_p^\text{min} = \phi$)
\begin{align}
    \begin{split}
        r'_\text{quad} &= 1 + \dot h^\text{cart}_\text{quad}(s, \dot s) +  \gamma  h^\text{cart}_\text{quad}(s) \\
           &= 1 + \delta_a((\phi^2 - \theta^2)-2\theta_p\omega_p), \delta_a \in \ \mathbb{R}_{>0}
    \end{split}
    \label{eq:cart-rew1}
\end{align}
\noindent \textbf{Results:} Fig. \ref{fig:cp1} shows energy expended till stabilisation vs initial angle for a range of angles. As can be read, the performance of policy using vanilla reward is energy-expensive compared to that of with $r^\text{BF}$. For one unit of energy spent by the vanilla policy, $\pi^\text{BF}_\text{exp}$ policy spends 0.79 units, $\pi^\text{BF}_\text{quad}$ policy spends 0.59 units. Fig. \ref{fig:Cartpole2} describes the control performance for the same initial state. In contrast to $\pi^\text{BF}_\text{quad}$ policy, the Vanilla policy exhibits chaotic behaviour with no convergence to the $\theta_p=0$ value. Thus our $\pi^\text{BF}_\text{quad}$ policy maintains safety during deployment. In Fig. \ref{fig:cp1} multiple barrier refers to $r^\text{BF}$ constructed on $\omega_p$ and $v_c$ along with $\theta_p$.

\vspace{-0.15cm}
\subsection{MuJoCo Walker Environments} \label{mujoco_walker}
\noindent \textbf{Reward Shaping:} For walkers like Half-Cheetah \cite{c3}, Ant \cite{c4}, and Humanoid \cite{c2}, the task is to learn to run as fast as possible. We use the same $r^\text{BF}$ formulation for all these environments as essentially they all have a similar state space $\mathcal{S}$ containing $x^w_{pos}$, $\Theta^w$ and $\Omega^w$, where $\Theta^w$ and $\Omega^w$ are the set of agent's joint angles and angular velocities, respectively. The vanilla reward $r$ given by (\ref{eq:walk-rew1}) has multiple terms that guide the agent to move forward ($r_\text{forward}$) without falling ($r_\text{health}$) and minimizing the contact forces ($r_\text{contact}$).
\begin{align}
    r = r_\text{health} + r_\text{forward} + r_\text{contact}.
    \label{eq:walk-rew1}
\end{align}
Since the task involves running, we seek to constrain the agent's joint angles $\theta_l\in\Theta^w$ within a range of angles $(\theta_l^\text{min}, \theta_l^\text{max})$ $\forall l\in\mathcal{L}$. In this case, $\mathcal{L}$ is the set of all joint angles. Thus, $\theta_l$ is analogous to $s_l$. These bounds correspond to the range of angles within which a joint can safely turn, violating which could lead to damage to the actuators or collision with other parts of the body. These general bounds are specified in the robot description and do not require additional domain knowledge. The optimal parameters $\bm{\delta}$ for \eqref{eq:quad-cbf}-\eqref{eq:exp-cbf} were found by performing a grid search.
 
\noindent \textbf{Metrics:} The metrics used to evaluate the policy performance are ($i$) \textit{Actuation Coefficient}, computed from the episodic energy-velocity curve (Fig. \ref{fig:enrg-vel}), ($ii$) \textit{Training Speed}, computed from the velocity-timestep curve (Fig. \ref{fig:vel-train}). All the experiments were repeated for ten random seeds. 

 The energy $\mathcal{E}_w$ used by a walker $w$ in a time-step is calculated by 
 \vspace{-0.3cm}
 \begin{align}
    \mathcal{E}_w = \sum_{j\in\mathcal{J}_w} \Delta \theta_j \cdot \tau_j \label{eq:agent-energy}
\end{align}
\vspace{-0.1cm}
where $\mathcal{J}_w$ is the set of all joints of the walker, $\tau_j$ is the torque exerted by the agent on the joint $j$, and $\Delta \theta_j$ is the change in the joint angle. The episodic energy is the sum of $\mathcal{E}_w$ over an episode. Since the task of these environments is to maximise the velocity $v_w$, we define the \textit{Actuation Coefficient} (given in Table. \ref{tab:main-res}) as 
 \vspace{-0.4cm}
 \begin{align}
    \text{(Actuation Coefficient)}_w = \frac{\mathcal{E}_w^\text{episodic}}{\left(v^\text{mean}_w\right)^2} \label{eq:act-coef}
\end{align}
\vspace{-0.1cm}
This term represents the energy expended or the effort made by the agent to achieve a certain kinetic energy. 

\noindent \textbf{Results:} Table \ref{tab:main-res} shows that for Humanoid, $\pi^\text{BF}_\text{exp}$ policy takes only about 49\% actuation energy to achieve the same kinetic energy as the vanilla policy. Fig. \ref{fig:enrg-vel} (Left) shows that the $\pi^\text{BF}$ policies reach a higher maximum velocity for the same training time-steps. Fig. \ref{fig:vel-train} (Left) shows that $\pi^\text{BF}_\text{exp}$ policy converges to a higher velocity in much fewer training time-steps. Table \ref{tab:main-res} shows that $\pi^\text{BF}_\text{exp}$ policy converges 1.56 times faster, while $\pi^\text{BF}_\text{quad}$ policy is no worse than vanilla policy. Continuing the trends from Table \ref{tab:main-res}, we see that the $\pi^\text{BF}$ policies on Ant and Half-cheetah share similarities with the results in the Humanoid environment. These similarities include reduced kinetic energy, quicker convergence, and increased agent velocity within the same time-step frame.

\section{SIM-TO-REAL ON HARDWARE} 

\begin{figure*}[t]
        \centering
        \begin{minipage}[b]{0.33\textwidth}
            \centering
            \includegraphics[width=\textwidth]{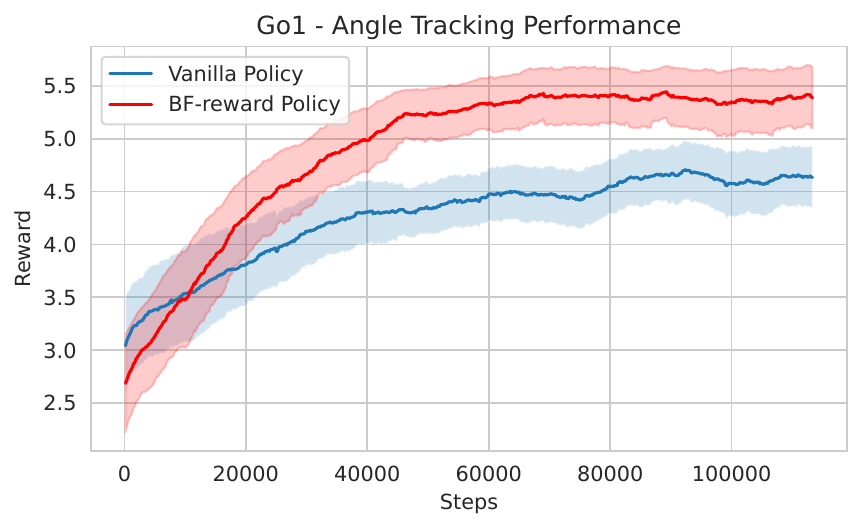}
        \end{minipage}
        \hspace{-1em}
        \begin{minipage}[b]{0.33\textwidth}
            \centering
            \includegraphics[width=\textwidth]{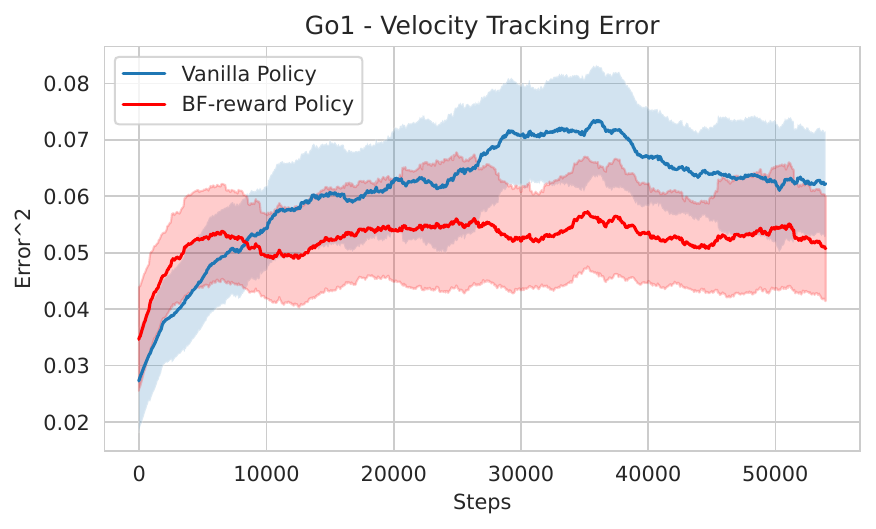}
        \end{minipage}
        \hspace{-1em}
        \begin{minipage}[b]{0.33\textwidth}
            \centering
            \includegraphics[width=\textwidth]{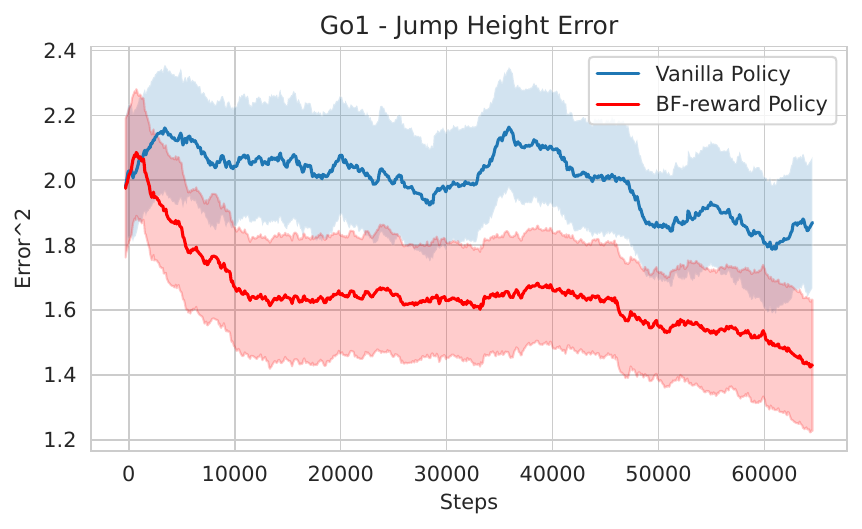}
        \end{minipage}    
        
        \begin{minipage}[b]{0.33\textwidth}
            \centering
            \includegraphics[width=\textwidth]{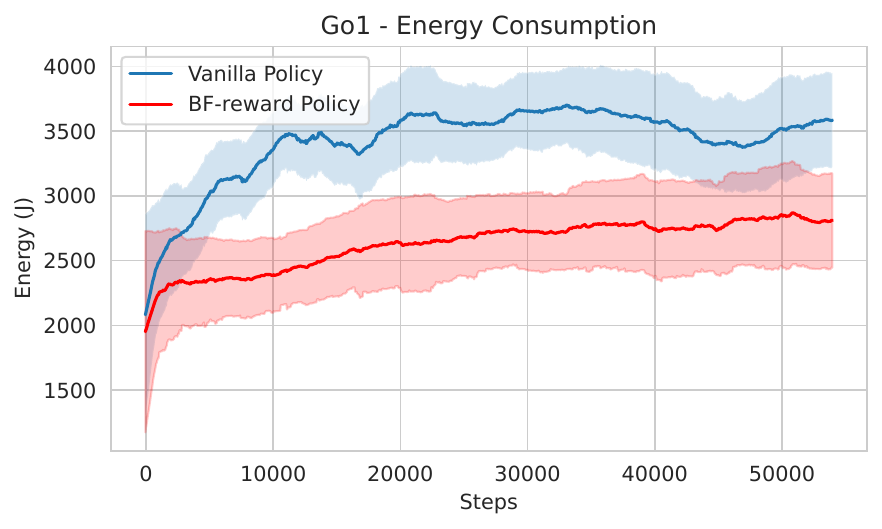}
        \end{minipage}
        \hspace{-1em}
        \begin{minipage}[b]{0.33\textwidth}
            \centering
            \includegraphics[width=\textwidth]{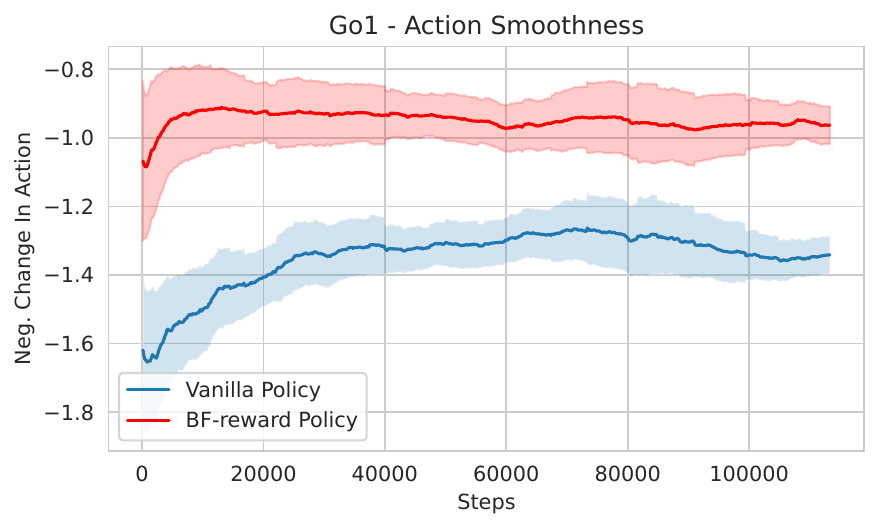}
        \end{minipage}
        \hspace{-1em}
        \begin{minipage}[b]{0.33\textwidth}
            \centering
            \includegraphics[width=\textwidth]{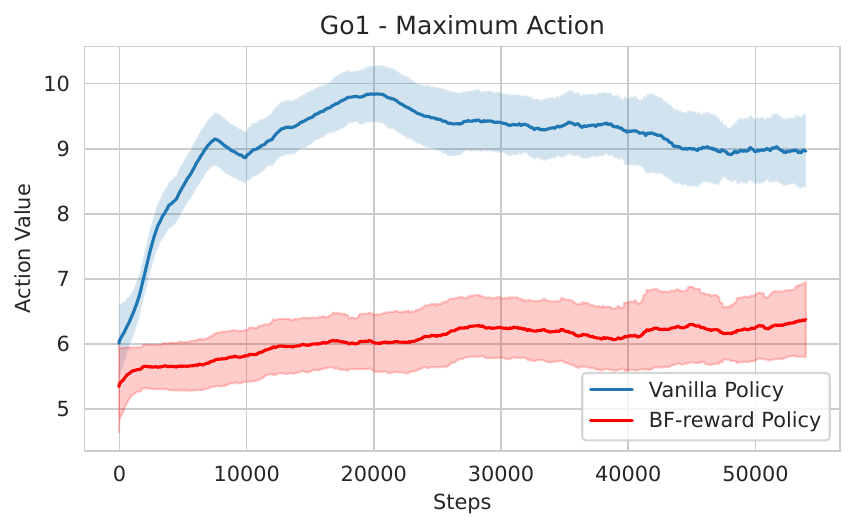}
        \end{minipage}
        \caption{Training plots for the Go1 robot in the Issac Gym simulation environment. The top row show that our policy $\pi^\text{BF}$ trained using $r^\text{BF}_\text{exp}$ outperforms the policy trained without it (vanilla) in all three tasks - Velocity Tracking, Angle Tracking, and Jumping. The angle tracking reward is inversely proportional to the error in angle. Note that for both the policies we are comparing the same vanilla reward in the plot. Action smoothness is negative of change in action. The bottom row provides insights about the safety and efficiency of $\pi^\text{BF}$.}
        \label{fig:go1-figs}
        \vspace{0.3cm}
        
        \begin{minipage}[b]{0.18\textwidth}
            \centering
            \includegraphics[width=\textwidth]{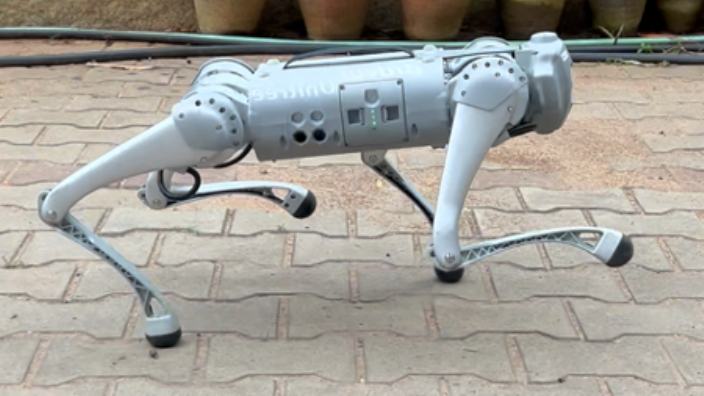}
        \end{minipage}
        \begin{minipage}[b]{0.18\textwidth}
            \centering
            \includegraphics[width=\textwidth]{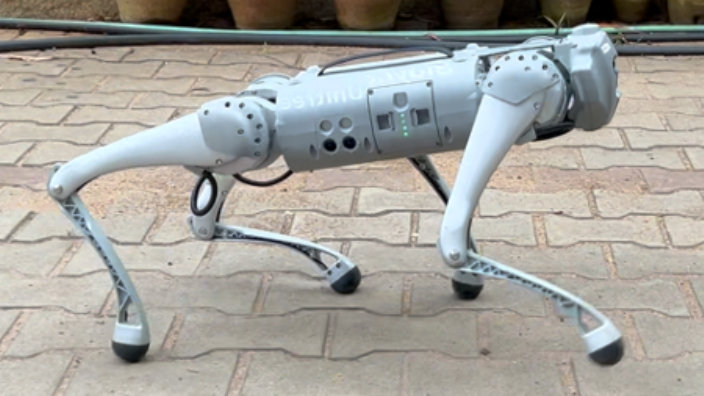}
        \end{minipage}
        \begin{minipage}[b]{0.18\textwidth}
            \centering
            \includegraphics[width=\textwidth]{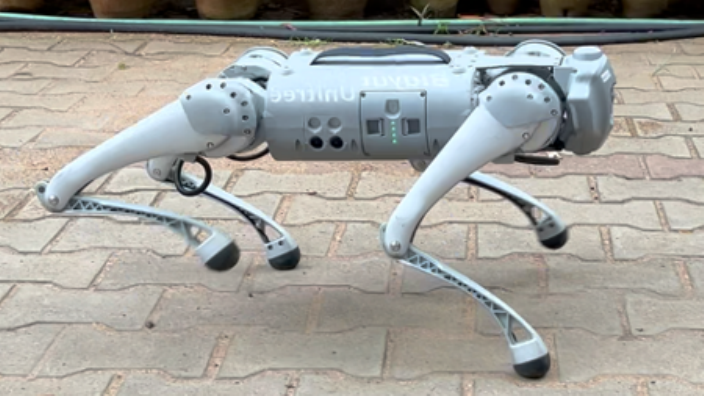}
        \end{minipage}
        \begin{minipage}[b]{0.18\textwidth}
            \centering
            \includegraphics[width=\textwidth]{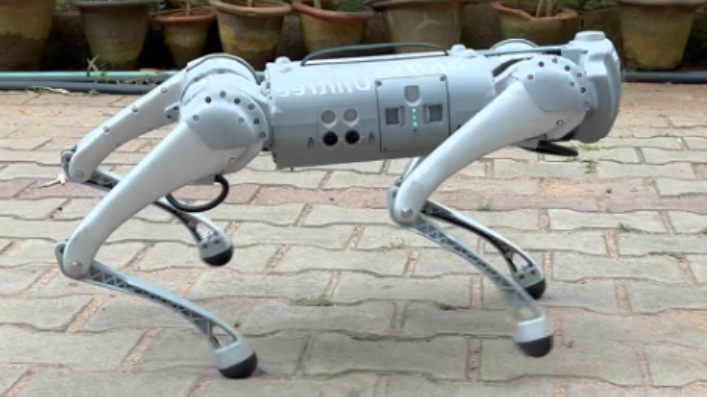}
        \end{minipage}
        \begin{minipage}[b]{0.18\textwidth}
            \centering
            \includegraphics[width=\textwidth]{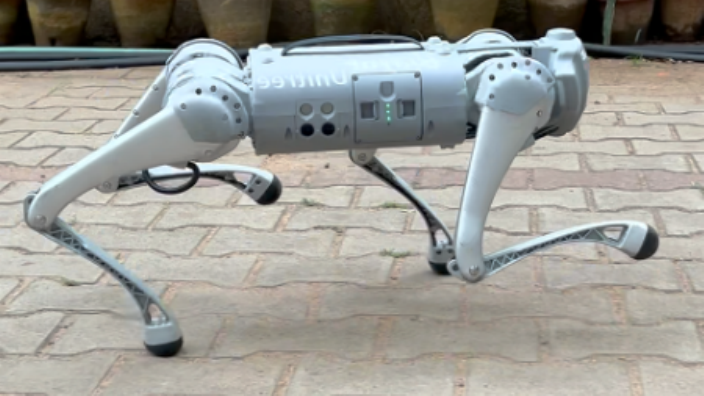}
        \end{minipage}
        
        \vspace{0.1cm}
        
        \begin{minipage}[b]{0.18\textwidth}
            \centering
            \includegraphics[width=\textwidth]{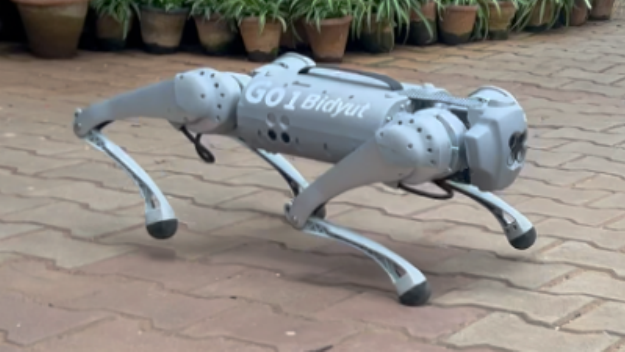}
        \end{minipage}
        \begin{minipage}[b]{0.18\textwidth}
            \centering
            \includegraphics[width=\textwidth]{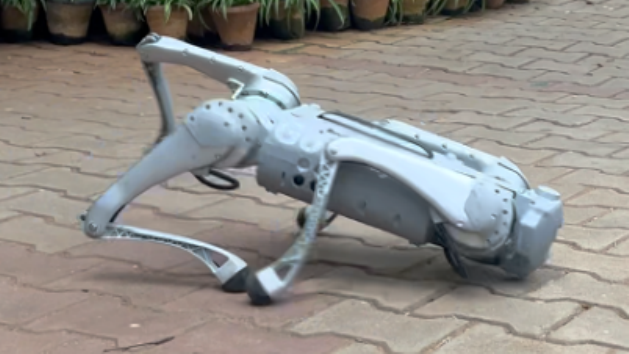}
        \end{minipage}
        \begin{minipage}[b]{0.18\textwidth}
            \centering
            \includegraphics[width=\textwidth]{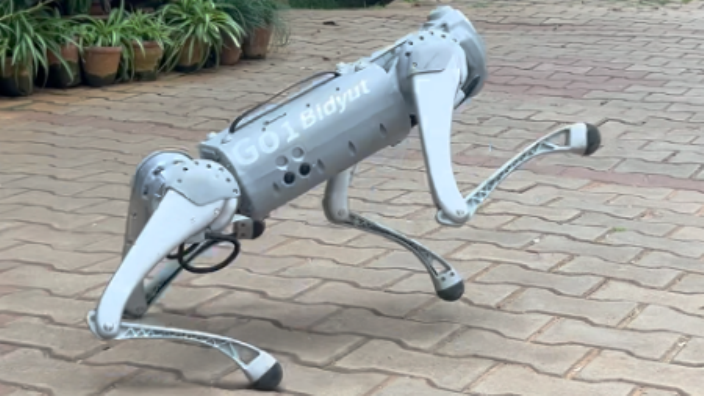}
        \end{minipage}
        \begin{minipage}[b]{0.18\textwidth}
            \centering
            \includegraphics[width=\textwidth]{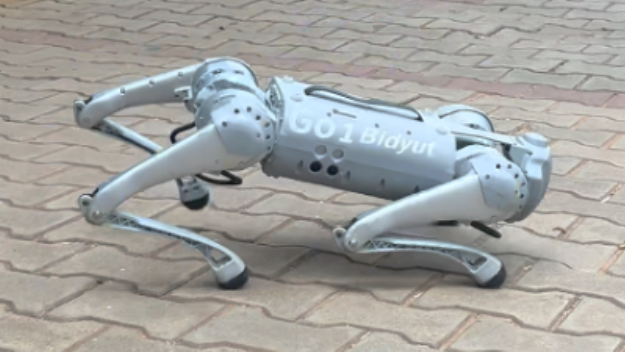}
        \end{minipage}
        \begin{minipage}[b]{0.18\textwidth}
            \centering
            \includegraphics[width=\textwidth]{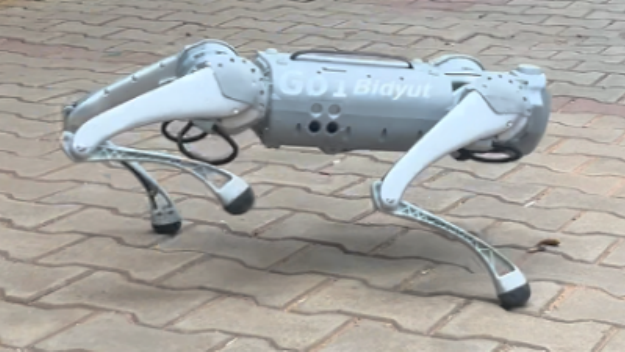}
        \end{minipage}

        \caption{Top row shows a sample run for the velocity tracking task using  $\pi^\text{BF}$ policy trained for 2.5k episodes. Bottom row shows the vanilla policy performance, trained for 5k episodes. Frames are 0.4 sec apart. The video can be found here: \url{http://www.stochlab.com/redirects/rewardshaping2023.html}.}
        \vspace{-0.3cm}
        \label{fig:go1-real}
        
    \end{figure*}

\subsection{Implementation Details}
\noindent \textbf{Hardware:} We deploy our trained policies on the Unitree Go1 robot \cite{unitree2022website}, a quadruped with 12 DOFs. We rely on the SDK provided by \cite{margolis2022walktheseways} to communicate between our code and the low-level control SDK provided by Unitree. The control frequency is 50Hz for both simulation and hardware.

\noindent \textbf{Simulation and Training:}  We use an open source implementation of the Go1 in the Isaac Gym simulator \cite{makoviychuk2021isaac}. We train the policies for 2500 episodes on 4000 parallel environments using PPO \cite{schulman2017proximal} with both the Vanilla Reward and $r^\text{BF}_\text{exp}$ constructed using $h_\text{exp}$ (\ref{eq:exp-cbf}), having same formulations as given in \ref{mujoco_walker} for other waslkers. The joint angle bounds were taken from the physical specifications of the robot.

\noindent \textbf{Domain Randomization:} To enhance sim-to-real transfer, we train a policy that remains robust across variations in robot attributes such as body mass, motor strength, joint position calibration, ground friction, restitution, and gravity orientation and magnitude by varying these quantities.

\subsection{Simulation and Training Results}
\begin{figure}[tp]
    \centering
      \begin{minipage}[b]{\textwidth}
  {\includegraphics[width = 0.24\textwidth]{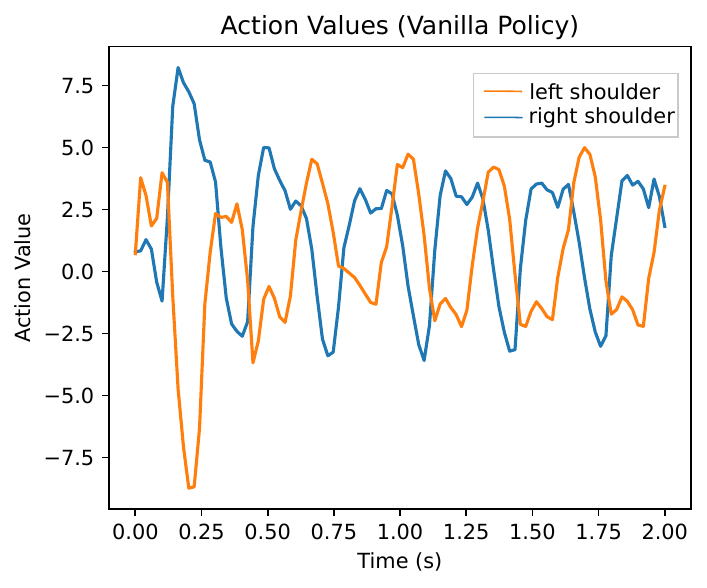}}
  {\includegraphics[width = 0.24\textwidth]{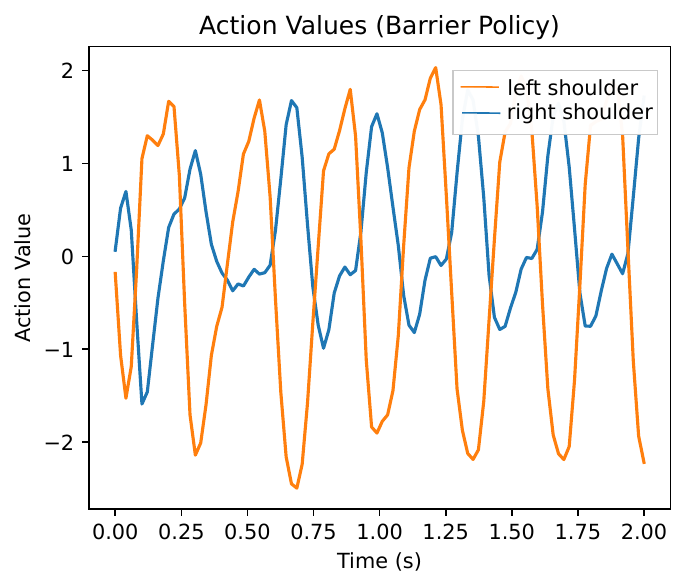}}  
  \end{minipage} 
    
    \caption{Action values for two front shoulder flexion joints of Go1 for Vanilla Policy (left) and $\pi^\text{BF}$ Policy (right) for a sample run of the velocity tracking task. The action values represent angle targets that are later clipped to a suitable range. It can be seen that the actions for $\pi^\text{BF}$ are smoother, rhythmic and have lower action value.}
    \vspace{-0.7cm}
    \label{fig:go1-anglerun}
\end{figure}
The Vanilla reward is comprised of three components: target velocity tracking, target angle tracking, and target height jump. As depicted in Fig. \ref{fig:go1-figs} (Top Row), our proposed policy $\pi^\text{BF}$ trained with the $r^\text{BF}_\text{exp}$ shaping term added to the vanilla reward consistently outperforms the Vanilla policy across all three tasks. In Fig. \ref{fig:go1-figs} (Bottom Row), it's evident that $\pi^\text{BF}$ utilizes only 78\% of the energy compared to Vanilla, as calculated using \eqref{eq:agent-energy}. Furthermore, $\pi^\text{BF}$ exhibits improved action smoothness (negative of change in action) and significantly lower maximum action values, thereby enhancing safety for deployment.


\subsection{Hardware Results}
\noindent\textbf{Vanilla Policy:} The Vanilla policy (Fig. \ref{fig:go1-real} Bottom) resulted in unstable movements and limb instability, particularly affecting the front right limb during basic commands like forward motion and re-direction (Fig. \ref{fig:go1-real}). This led to a fall, from which the robot autonomously recovered but at the cost of posing practical deployment risks. The robot's joint movements lacked coordination and efficiency, frequently deviating from the intended path and causing the front hinge joints to approach dangerously close to the ground.


\noindent\textbf{$\pi^\text{BF}$ Policy:} The policy trained with the new reward significantly enhanced the Go1 robot's performance, with improved balanced and coordinated movements despite having half the training duration as the Vanilla policy (Fig. \ref{fig:go1-real} Top). It flawlessly followed RC controller commands, ensuring safety without falls or risky behaviour. The advantages of the $\pi^\text{BF}$ policy are further affirmed by Fig. \ref{fig:go1-anglerun} where $\pi^\text{BF}$ policy leads to more consistent and rhythmic action values. This characteristic enhances safety for the actuators and contributes to increased task efficiency.


\section{CONCLUSIONS}
In this paper we propose barrier function (BF) inspired reward shaping, a safety-oriented, easy-to-implement reward shaping formulation for robotic platforms. This approach is based on theoretical principles of safety provided by barrier functions. The shaping term aims to encourage agents to remain within predefined safe states during training. This enhances training efficiency and ensures safer exploration. To illustrate our formulation process, we proposed two barrier functions: \textit{exponential} and \textit{quadratic}.

While prior works, e.g. by Cheng et al. \cite{Cheng_Orosz_Murray_Burdick_2019} achieved high training efficiency and state safety, their framework needs the system dynamics model. In contrast, our method eliminates this need, thus being easy to implement in complex environments. However, a limitation of our study is that we have only tested with barrier functions of joint angles. Further investigation into other quantities such as joint angular velocities could offer a more comprehensive understanding of our reward's effectiveness.


We employed the Unitree Go1 robot \cite{unitree2022website} as our hardware platform for sim-to-real experiments. Our reward-shaping methodology emerged superior through comparative analysis with \cite{margolis2022walktheseways}, revealing smoother, more rhythmic control dynamics. The results indicate that our formulation is an easy way to introduce safety and efficiency in RL training.

\bibliographystyle{IEEEtran}
\bibliography{references}

\end{document}